\pdfoutput=1
\documentclass[a4paper]{article}

\usepackage{graphicx}

\usepackage{INTERSPEECH2022}
\usepackage{subfig}
\usepackage{amssymb}
\usepackage{multirow}
\usepackage{bbding}
\usepackage{float}
\title{SF-DST: Few-Shot Self-Feeding Reading Comprehension Dialogue State Tracking with Auxiliary Task}
\name{Jihyun Lee$^1$, Gary Geunbae Lee$^1$,$^2$}
\address{
  $^1$Graduate School of Artificial Intelligence, POSTECH, Republic of Korea\\
  $^2$Department of Computer Science and Engineering, POSTECH, Republic of Korea}
\email{jihyunlee@postech.ac.kr, gblee@postech.ac.kr}

\begin{document}

\maketitle
\begin{abstract}
Few-shot dialogue state tracking (DST) model tracks user requests in dialogue with reliable accuracy even with a small amount of data. In this paper, we introduce an ontology-free few-shot DST with self-feeding belief state input. The self-feeding belief state input increases the accuracy in multi-turn dialogue by summarizing previous dialogue. Also, we newly developed a slot-gate auxiliary task. This new auxiliary task helps classify whether a slot is mentioned in the dialogue. Our model achieved the best score in a few-shot setting for four domains on multiWOZ 2.0.

\end{abstract}

\noindent\textbf{Index Terms}: multi-domain dialogue systems, dialogue state
tracking, belief tracking, reading comprehension, self-feeding

\section{Introduction}
\begin{figure}[t]
  \centering
  \includegraphics[width=1.0\linewidth]{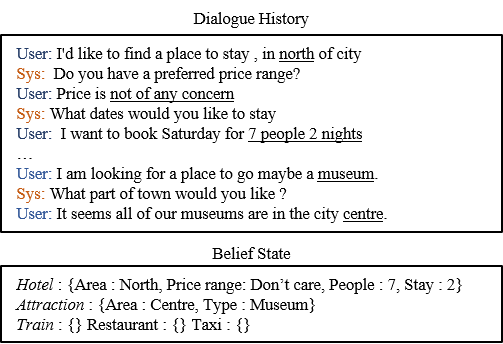}
  \caption{Example of multi-domain TOD dataset and its belief state. The dialogue state tracker (DST) keeps track of user’s requests and predicts value of slot.}
  \label{fig:dialogue_example}
\end{figure}

Task-Oriented Dialogue (TOD) system conducts a conversation with a specific purpose and is increasingly necessary due to the emergence of artificial-intelligence speakers and virtual personal assistants. In general, a TOD system is composed of three main sections: a dialogue state tracking (DST) module to track the user's purpose, a dialogue policy module (POL) to choose system actions like \textit{API calling} or \textit{ending conversation}, and a natural-language generation (NLG) module to produce a response to the user \cite{young2013pomdp}. The DST system is a key component of these three parts since it generates a belief state that contains information about the user's purpose. The belief state is often represented as slot value pairs. For example, in Figure~\ref{fig:dialogue_example}, the belief state has hotel information, which is needed to achieve user's purpose for hotel reservation.

Although DST is essential in a TOD system, labeling the DST dataset is costly. Some authors have tried to train DST using only limited data (few/zero-shot DST) to solve this problem. One promising way is to adopt reading comprehension (RC) in DST  \cite{gao2020machine, li2021zeroshot}. The RC task aims to answer the question by understanding the passage. RC and DST have a similar goal: DST aims to find the value (answer) of slot (question) by understanding the dialogue (passage). In this approach, researchers design questions (e.g., \textit{Where is hotel area that the user wants?}) for slots (e.g., \textit{hotel.area}) in advance, and at each turn of dialogue, the model reads the dialogue and answers the questions. These predicted answers become belief state. The first research \cite{gao2020machine} that adopted RC in a DST divided slots into multiple-choice and span-prediction types and showed knowledge transfer ability of natural language questions. However, their model requires an ontology data that contains pre-defined values for each slots. This has low scalability to new domain and values. To overcome the limitation,  \cite{li2021zeroshot} proposed an ontology-free model by answering questions generatively and could flexibly predict the unseen domain and values. However, both \cite{gao2020machine} and \cite{li2021zeroshot} require additional external data for pre-training their models. Also, their models have difficulty in classifying whether a slot is mentioned in the dialogue and this problem was the main reason for the accuracy drop.

The use of an auxiliary task is another approach of a few-shot DST. The named-entity recognition task was combined with the DST task to reduce the number of network parameters and increase generalization ability across the domain \cite{rastogi2018multi,lee2021improving}. The language modeling task was also used with the main DST task, and the combination increased the accuracy in a long context dialogue  \cite{quan2020modeling}.

In this study, we introduce a few-shot reading comprehension DST with a \textbf{S}elf-\textbf{F}eeding approach (SF-DST). We used a text-to-text structure for generative, ontology-free DST and designed a self-feeding belief state input to summarize the previous turn. Applying self-feeding belief state is the first attempt in the RC format DST. Furthermore, we developed a slot-gate auxiliary task which helps to classify whether a slot is mentioned in the dialogue.

Our model achieved the new best accuracy in a few-shot experiment for four domains on MultiWOZ 2.0  \cite{budzianowski2018large}, and achieved close to the current state-of-the-art in a supervised setting on MultiWOZ 2.1 \cite{eric2019multiwoz}. In analysis, we investigated the effect of self-feeding belief state input and auxiliary task in various few-shot settings. To summarize our approach and contributions:

\begin{itemize}
\item We propose \textbf{S}elf-\textbf{F}eeding reading comprehension \textbf{DST} (SF-DST), an ontology-free few-shot model to track belief state. SF-DST has a text-to-text structure and includes self-feeding belief state input. We showed that the belief state helps understand multi-turn dialogue by summarizing previous turns.

\item We introduce a novel auxiliary task inspired by slot-gate in extractive DST. This new auxiliary task helps distinguish whether a slot is mentioned in the dialogue. We carefully analyzed the effect of the auxiliary task.

\item In a few-shot setting, where 1\% to 10\% data is available, SF-DST achieved higher accuracy than the previous methods for four domains on MultiWOZ 2.0. In a supervised setting, SF-DST was close to the current state-of-the-art on MulitiWOZ 2.1.
\end{itemize}
\section{Methods}
\begin{figure}[t]
  \centering
  \includegraphics[width=\linewidth]{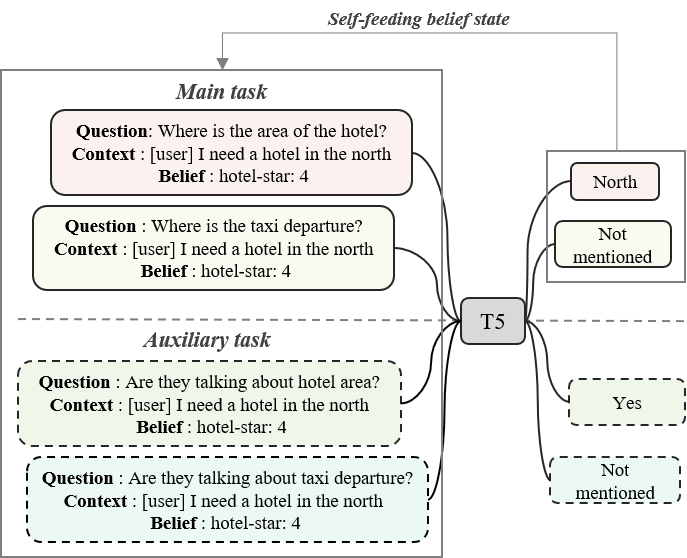}
  \caption{The proposed SF-DST model's architecture. Our model has a text-to-text structure and uses the reading comprehension method. The model receives questions, dialogue history, and predicted previous belief state as inputs (self-feeding). In addition to the DST task, the model is trained on a slot-gate auxiliary task (marked with a dashed line).}
  \label{fig:model_figure}
\end{figure}

\subsection{Problem statement}
Conversation $C$ for time step $t$ is denoted as $C_t = \{x_1,y_1,...,y_{t-1},x_t\} $, where $x_{t}$  means user utterance and $y_{t}$ means system utterance. The belief state $B_{t}$ at turn ${t}$ is composed of slot $s\in S$ and value $v \in V$. $V$ includes \textit{don't care} and \textit{not mentioned} values and notation $v_s$ means value for slot $s$. Question set ${Q}$ consists of ${q_s}$ which is predefined before training (e.g., $s$: \textit{attraction.name}, ${q_s}$: \textit{What is the attraction name?}). Auxiliary set ${A}$ consists of ${a_{s}}$, and '\textit{Are they talking about [slot]?}' is the inquiry form. ${a_{s}}$ is also predefined before training (e.g., $s$: \textit{attraction.name}, ${a_{s}}$: \textit{Are they talking about attraction name?}).

\subsection{SF-DST}
SF-DST has text-to-text structure for generative, ontology-free DST  (Figure~\ref{fig:model_figure}). The input value consists of dialogue history $C_t$, corresponding question for slot $q_s$, and previous belief state $B_{t-1}$. The model answers the question for all slots at each dialogue turn $t$,  and the predicted answers $v_s$ become $B_t$. We did not use a gold belief state as input in both training and inference time; instead, we designed a self-feeding belief state method in which the predicted belief state from the previous turn becomes the current turn's input belief state. We separate user and system utterances by using \textit{[user]} and \textit{[sys]}, and add index words \textit{Context}, \textit{Question} and \textit{Belief} to distinguish each input part:

\begin{equation}
  v_s = seq2seq(C_t, q_s, B_{t-1}).
  \label{input}
\end{equation}
We use negative log-likelihood as a loss function given $C_t$, $q_s$ and $B_{t-1}$ as

\begin{equation}
  L_{belief} = -\sum_{i=1}^{n} \log p(v_i|C_t,q_i,B_{t-1}),
  \label{belief_loss}
\end{equation}
where $n$ denotes the total number of slots.

\subsection{Auxiliary task}
Some extractive DSTs use a two-step system. These systems first classify whether the slot is mentioned in dialogue, and if classified as mentioned, then finds the answer span from dialogues \cite{WuTradeDST2019, chao2019bert, heck2020trippy,kumar2020ma}. This classification module is called slot-gate. By splitting the DST task into two models, this strategy can lower the strain on each model. However, this cascading approach risks error propagation and requires a relatively long inference time. Instead of a cascading strategy, we directly answered a question and trained a slot-gate task as an auxiliary task. The auxiliary task is trained as a question-answering form (Figure ~\ref{fig:model_figure}) and '\textit{Are they talking about [slot]?}' is the inquiry format. The answer is \textit{Yes} if the slot value is in belief state $B_t$, and \textit{not mentioned} otherwise; i.e., the main DST question aims to generate a specific value, whereas the auxiliary question aims to classify the slot's mention in the dialogue. The auxiliary task uses the context $C_t$, previous belief state $B_{t-1}$ and auxiliary question $a_s$ as input which has the same form as (1) except the slot question, and uses loss function (\ref{belief_loss}). To train the auxiliary task with the main DST task, our model uses a joint loss function with hyperparameter $a$
\begin{equation}
  L = L_{belief} + aL_{aux}.
  \label{loss}
\end{equation}
We set $a$ empirically to 0.7.

\begin{table}[b]
  \caption{Joint goal accuracy on MultiWOZ 2.1 in a supervised setting. Models focused on a few/zero shot are marked with $\dagger$.}
  \label{tab:full_result}
  \centering
  \begin{tabular}{lcccc}\hline
    Model&JGA [\%]&Ontology&Type\\
    \midrule
    TRADE $\dagger$ & 46.00 & & G\\
    STARC $\dagger$ & 49.48   & need & C+S\\
    DSTQA $\dagger$  & 51.17  & need & C+S\\
    DS-DST & 51.21 &   & C+S\\
    GPT2QA $\dagger$ & 52.58 & & G\\
    SST-2 & 55.23 &  need & C\\
    TripPy & 55.29 &  & S\\
    FPDSC$_{turn}$ & \textbf{57.88} & need & C\\

    \midrule
    SF-DST (ours) &  \textbf{56.96} & & G \\
  \bottomrule
\end{tabular}
\end{table}
\section{Experiment}

\begin{table*}[t]
\caption{Domain joint goal accuracy [\%] on MultiWOZ 2.0. We use 2.0 version to compare with other models. Models that require ontology are marked with $\dagger$.}
\label{tab:few_result}
\centering
\resizebox{\textwidth}{!}{%
\begin{tabular}{lccccccccccccccc}
\hline
\multirow{2}{*}{Model} & \multicolumn{3}{c}{Hotel} & \multicolumn{3}{c}{Restaurant} & \multicolumn{3}{c}{Attraction} & \multicolumn{3}{c}{Train} & \multicolumn{3}{c}{Taxi} \\ \cline{2-16} 
                       & 1\%     & 5\%    & 10\%   & 1\%      & 5\%      & 10\%     & 1\%      & 5\%      & 10\%     & 1\%     & 5\%    & 10\%   & 1\%    & 5\%    & 10\%   \\ \hline
TRADE                  & 19.73   & 37.45  & 41.42  & 42.42    & 55.70    & 60.94    & 35.88    & 57.55    & 63.12    & 59.83   & 69.27  & 71.11  & 63.81  & 66.58  & 70.19  \\
DSTQA $\dagger$        & N/A     & 50.18  & 53.68  & N/A      & 58.95    & 64.51    & N/A      & 70.47    & 71.60    & N/A     & 70.35  & 74.50  & N/A    & 70.90  & 74.19  \\
STARC $\dagger$        & 45.91   & 52.59  & 57.37  & 51.65    & 60.49    & 64.66    & 40.39    & 65.34    & 66.27    & 65.67   & 74.11  & 75.08  & \textbf{72.58}  & \textbf{75.35}  & \textbf{79.61}  \\ \hline
SF-DST                 & \textbf{54.15}   & \textbf{58.61}  & \textbf{59.71}  & \textbf{57.28}    & \textbf{68.92}    & \textbf{70.14}    & \textbf{61.24}
                       & \textbf{76.69}   & \textbf{79.35}  & \textbf{71.60}   & \textbf{76.05}  & \textbf{78.25}      & 65.74      & 67.48      & 72.06      \\ \hline
\end{tabular}%
}
\end{table*}

We performed experiments on MultiWOZ 2.0 and MultiWOZ 2.1 dataset, which are multi-domain TOD datasets collected using a 'wizard of oz' setting. MultiWOZ 2.1 is a clean and accurate version of MultiWOZ 2.0. Both have seven domains (\textit{Hotel, Restaurant, Attraction, Train, Taxi, Hospital, and Police}) and contain about 8,000 dialogues. We excluded the \textit{Hospital} and \textit{Police} domains during training, because they are only included in training data. We use joint goal accuracy (JGA) to evaluate our model's accuracy: If all slot and value pairs in turn are correct, the turn is counted as correct, and joint goal accuracy is the average value of all turns.

In Section \ref{supervised}, we evaluate our model in a supervised setting. In this setting, we use all training dataset and compare with other DSTs. In section \ref{fewshot}, we evaluate our model in a few-shot setting, where our primary focus lies. We report the parameter size of the baselines and detailed implementation in the Appendix.

\subsection{DST with supervised setting}
\label{supervised}
We evaluated our model in a commonly-used supervised setting to compare with other DST models. In comparison, we included few/zero-shot DST baselines---TRADE \cite{trade-dst}, STARC \cite{gao2020machine}, DSTQA \cite{zhou2020multidomain}, GPT2QA\footnote{The paper did not named the model. We assign temporary name to simplify comparison.} \cite{li2021zeroshot}---and other general DST models, including DS-DST \cite{zhang2019find}, SST-2 \cite{chen2020schema}, TripPy \cite{heck2020trippy}, and FPDSC \cite{zhou2021dialogue}. In addition to JGA,  we also report ontology usage and answer prediction type. We divide answer predict type into  classification (C), span prediction (S), and generative (G) following \cite{li2021zeroshot}.
SF-DST achieved the highest accuracy by a wide margin compared to other few/zero-shot DST (Table~\ref{tab:full_result}). Our accuracy was lower than the best score of the classification-type method. However, these models need fixed ontology and find answers by classifying the value in ontology \cite{gao2020machine, zhou2020multidomain, zhang2019find, chen2020schema, zhou2021dialogue}. Fixed ontology is hard to obtain in the real world and cannot adapt to frequently changing values. In contrast, our model is ontology-free and generates the answer, which is competitive in the real world.

\subsection{DST with few-shot data setting}
\label{fewshot}

\label{fewshot_domain}

\subsubsection{Domain across knowledge transfer in the few-shot setting}

To investigate the knowledge transfer ability across the domain, we pre-trained our model on four domains and fine-tuned it with the target domain. We used domain JGA, which measures the JGA focused on the targeted domain \cite{campagna2020zeroshot}. SF-DST exceeded the previous best score in four of the five domains compared to other few-shot DST results, including TRADE \cite{trade-dst}, DSTQA \cite{zhou2020multidomain}, and STARC \cite{gao2020machine} (Table~\ref{tab:few_result}). This result demonstrates that our model can adapt to a new domain using only limited data by transferring knowledge from other domains. However, SF-DST showed lower accuracy than ontology-based models (marked as $\dagger$ in table) in the taxi domain. The taxi domain is generally mentioned at the end of the dialogue \cite{zhou2020multidomain}, so even with the same size of data, the chance of training taxi domain is much lower than other domains. Under this condition, the ontology classification method has an advantage over the generative method.

\subsubsection{Comparison with few-shot TOD systems}
\label{fewshot_all}
\begin{table}[h]
\caption{Joint goal accuracy [\%] on MultiWOZ 2.1 and use of external data. The result of SimpleTOD, MinTL, SOLOIST and PPTOD are referenced from \cite{su2021multitask}.}
\label{tab:few_tod}
\centering
\begin{tabular}{lcccc} \hline
\multirow{2}{*}{Model} &\multirow{2}{*}{External Data} &\multicolumn{3}{c}{Training Size (\%)}   \\ \cmidrule{3-5} 
                       & &  1       & 5       & 10        \\
\midrule
SimpleTOD              & No & 7.91    & 16.14   & 22.37      \\
MinTL                  & No & 9.25    & 21.28   & 30.32      \\
SOLOIST                & Yes & 13.21   & 26.53   & 32.42      \\
PPTOD$_{small}$        & Yes & 27.85   & 39.07   & 42.36 \\ 
\midrule
SF-DST                 & No & \textbf{28.35}  &\textbf{39.39}   & \textbf{44.60}    \\

\bottomrule
\end{tabular}
\end{table}

End-to-end TOD systems generally has DST, Policy, and NLG modules, and in the real world, our model could be used as a DST module of these systems. Therefore, we compared SF-DST with other DSTs in the TOD systems. We varied the training data rate as 1\%, 5\%, or 10\% and compared with SimpleTOD \cite{hosseiniasl2020simple}, MinTL \cite{lin2020mintl}, SOLOIST \cite{peng2021soloist} and PPTOD  \cite{su2021multitask}. SF-DST yielded the best accuracy in all few-shot settings (Table~\ref{tab:few_tod}). Our model does not rely on external data, so it can be simply implemented in existing TOD systems. From this result, we anticipate that our model can improve the TOD system as a plug-and-play DST module.

\section{Analysis}

\label{analysis}

\subsection{Ablation study}
\begin{table}[ht]
\caption{Ablation study of SF-DST, reporting few-shot (10\%) JGA on MultiWOZ2.1 data.}
\label{tab:ablation study}
\centering
  \begin{tabular}{lc}\hline
    Ablation& JGA [\%] \\
    \midrule
    SF-DST (this work) & 44.60 \\
    -- Self-feeding belief state & 42.31 \\
    -- Auxiliary task & 42.69 \\
    -- Self-feeding belief state + Gold belief state & 38.55 \\
  \bottomrule
\end{tabular}
\end{table}

We perform an ablation study to investigate which component contributes to accuracy in a few-shot environment (10\%). We observe that both self-feeding belief state and auxiliary task are essential to increase the accuracy (Table~\ref{tab:ablation study}). Additionally, we trained the model with the gold belief state (\textit{+ Gold belief state} in table) instead of self-feeding belief state. The JGA showed a significant decrease (38.55\%) compared to training with a self-feeding belief state (44.60\%). When the gold belief state is given during the training, the model depends on the belief state rather than the conversation. This causes the performance degrades in the inference stage, where the gold belief state cannot be given.

\subsection{Analysis of self-feeding belief state input}

\begin{table}[h]
\caption{Analysis of self-feeding belief state input. We separate the dialogue by the previous turn length and evaluate turn JGA.}
\label{tab:analysis_belief}
\centering
\begin{tabular}{lcccc} \hline
\multirow{2}{*}{Model} & \multicolumn{3}{c}{Previous dialogue turns} \\ \cmidrule{2-4} 
                        & 1 to 3 & 4 to 6 & 7+   \\
\midrule

SF-DST   & 54.68  & 29.65 & 14.05 \\
-- belief state (ablation) & 52.16 & 27.42  & 15.05 \\

\bottomrule
\end{tabular}
\end{table}
Although accurate DST requires an understanding of the entire dialogue, it is challenging when the dialogue has many turns. In this experiment, we analyzed the effect of the belief state according to the conversation length in a multi-turn circumstance. We separated the dialogue into three classes by the number of previous turns: short-length dialogue (one to three turns), medium-length dialogue (four to six turns), and long-length dialogue (seven or more turns). We trained the model with and without belief state in a few-shot setting (10\%) and used the average of turn JGA. Our self-feeding belief state improved both short-length dialogue and medium-length dialogue (Table~\ref{tab:analysis_belief}). This means that the belief state, which summarizes previous conversation information, helped the model to understand the multi-turn conversation. However, the JGA decreased when the dialogue was extended (more than seven turns). As the dialogue progressed, the probability of error propagating from the previous belief state increased, so the accuracy dropped. Therefore, finding a self-feeding method that reduces error propagation is a worthy future goal.

\subsection{Error analysis and effect of auxiliary task}
\begin{table}[h]
\caption{Changed error and JGA rate by adopting auxiliary task on MultiWOZ 2.1. Upper triangle 3.48 means an error rate increases 3.48\% point.}
\label{tab:analysis_aux}
\centering
\begin{tabular}{lcccc} \hline
\multirow{2}{*}{Error Type} & \multicolumn{3}{c}{Training Size(\%)} \\ \cmidrule{2-4} 
                       & 1       & 5       & 10           \\
\midrule
Wrong    &$\triangle$3.48  &$\triangle$16.47  & $\triangle$8.97       \\
Ignore   &$\bigtriangledown$10.6      &$\bigtriangledown$22.21   &$\bigtriangledown$20.46  \\
Spurious &$\bigtriangledown$10.29     &$\bigtriangledown$1.17   &$\bigtriangledown$4.22    \\
\midrule\midrule
JGA &$\triangle$13.96  &$\triangle$12.04  & $\triangle$4.48\\ 
\bottomrule
\end{tabular}
\end{table}

To examine the effect of the slot-gate auxiliary task, we classified the errors as \textit{Wrong}, \textit{Ignore}, and \textit{Spurious} \cite{gao2020machine, li2021zeroshot}. \textit{Wrong} means that the model correctly predicts the existence of the answer but predicts the wrong value. \textit{Ignore} means that the answer exists, but the model ignores it. \textit{Spurious} means that the answer is not mentioned, but the model predicts some value. We trained our model with and without auxiliary tasks at various training sizes and calculated the changed error and JGA rate by adapting the auxiliary task (Table~\ref{tab:analysis_aux}). 
Overall, JGA was improved in all data settings. This result indicates that the auxiliary task helped to find accurate answers. In the case of error type, \textit{ignore}, and \textit{spurious} errors decreased; this result means that the auxiliary task was helpful to classify whether a slot is mentioned in the dialogue. However, the \textit{wrong} type error grows in all settings. Adopting the auxiliary task increases the number of attempts to find the answer in dialogue when the answer exists, and this causes the growth of the \textit{wrong} type error. Future work should find an auxiliary task that decreases all types of errors.

\subsection{Implicit answers and auxiliary task}
\begin{table}[h]
\caption{Changed JGA by adopting auxiliary task on MultiWOZ 2.1. Upper triangle 5.74 means a JGA increases 5.74\% point.}
\label{tab:analysis_five}
\centering
\begin{tabular}{lcccc} \hline
\multirow{2}{*}{Slot Type} & \multicolumn{3}{c}{Training Size(\%)} \\ \cmidrule{2-4} 
                       & 1       & 5       & 10           \\
\midrule

Explicit    &$\triangle$ 5.74  &$\triangle$ 1.13  &$\triangle$ 2.26      \\
Implicit &$\triangle$ 5.74 &$\triangle$ 1.70   &$\triangle$ 2.57 \\

\bottomrule
\end{tabular}
\end{table}

Finding the proper answer becomes increasingly challenging when the dialogue does not include an exact match. For example, assume the user said, "I want to find a place to see a movie." In that situation, even if the attraction type is not explicitly given, the model should infer the attraction type as theater. This implicit answer circumstance is common in the real world, and we experimented to determine whether our auxiliary question assists in such conditions. We chose ten slots\footnote{train.day, restaurant.area, hotel.star, attraction.area, hotel.stay, hotel.area, restaurant.day, hotel.people, hotel.day, restaurant.pricerange} with a high probability of exact matching answers (explicit slot) and ten slots\footnote{hotel.type, hotel.internet, hotel.parking, taxi.leaveat, attraction.name, taxi.departure, attraction.type, train.leaveat, taxi.destination, hotel.name} with a low probability of exact matching answers (implicit slot) \cite{gao2020machine}. In the case of explicit slots, 99.12 \% of the answers were exactly found in dialogue, compared to 70.34\% in implicit slots. We experimented in a few-shot setting and used the JGA of targeted slots. Our auxiliary task improved JGA in all training data set and slot types (Table~\ref{tab:analysis_five}). Asking whether the slot is mentioned helps find an answer even if there is no exact match exists.

\section{Conclusion}
This paper proposed a generative few-shot DST that has a reading comprehension approach. Our text-to-text model is ontology-free and does not use external data. As an input, we devised a self-feeding belief state and showed that summarized information of belief state is helpful for multiple turn dialogue. Also, we developed a slot-gate auxiliary task. This task reduces the \textit{ignore} and \textit{spurious} type errors. As a result, in a few-shot experiment, SF-DST was more accurate than the previous methods for four domains on MultiWOZ 2.0  and was close to the state-of-the-art in a supervised experiment on MultiWOZ 2.1. 

\section{Acknowledgements}
This work was supported by SAMSUNG Research, Samsung Electronics Co.,Ltd., and  also supported by Institute of Information \& communications Technology Planning \& Evaluation(IITP) grant funded by the Korea government(MSIT) (No.2021-0-00575, Development of Voicepishing Prevention Technology Based on Speech and Text Deep Learning)

\bibliographystyle{IEEEtran}
\bibliography{my}

\begin{thebibliography}{10}
\providecommand{\url}[1]{#1}
\csname url@samestyle\endcsname
\providecommand{\newblock}{\relax}
\providecommand{\bibinfo}[2]{#2}
\providecommand{\BIBentrySTDinterwordspacing}{\spaceskip=0pt\relax}
\providecommand{\BIBentryALTinterwordstretchfactor}{4}
\providecommand{\BIBentryALTinterwordspacing}{\spaceskip=\fontdimen2\font plus
\BIBentryALTinterwordstretchfactor\fontdimen3\font minus
  \fontdimen4\font\relax}
\providecommand{\BIBforeignlanguage}[2]{{%
\expandafter\ifx\csname l@#1\endcsname\relax
\typeout{** WARNING: IEEEtran.bst: No hyphenation pattern has been}%
\typeout{** loaded for the language `#1'. Using the pattern for}%
\typeout{** the default language instead.}%
\else
\language=\csname l@#1\endcsname
\fi
#2}}
\providecommand{\BIBdecl}{\relax}
\BIBdecl

\bibitem{young2013pomdp}
S.~Young, M.~Ga{\v{s}}i{\'c}, B.~Thomson, and J.~D. Williams, ``Pomdp-based
  statistical spoken dialog systems: A review,'' \emph{Proceedings of the
  IEEE}, vol. 101, no.~5, pp. 1160--1179, 2013.

\bibitem{gao2020machine}
S.~Gao, S.~Agarwal, T.~Chung, D.~Jin, and D.~Hakkani-Tur, ``From machine
  reading comprehension to dialogue state tracking: Bridging the gap,''
  \emph{arXiv preprint arXiv:2004.05827}, 2020.

\bibitem{li2021zeroshot}
S.~Li, J.~Cao, M.~Sridhar, H.~Zhu, S.-W. Li, W.~Hamza, and J.~McAuley,
  ``Zero-shot generalization in dialog state tracking through generative
  question answering,'' 2021.

\bibitem{rastogi2018multi}
A.~Rastogi, R.~Gupta, and D.~Hakkani-Tur, ``Multi-task learning for joint
  language understanding and dialogue state tracking,'' \emph{arXiv preprint
  arXiv:1811.05408}, 2018.

\bibitem{lee2021improving}
Y.~Lee, ``Improving end-to-end task-oriented dialog system with a simple
  auxiliary task,'' in \emph{Findings of the Association for Computational
  Linguistics: EMNLP 2021}, 2021, pp. 1296--1303.

\bibitem{quan2020modeling}
J.~Quan and D.~Xiong, ``Modeling long context for task-oriented dialogue state
  generation,'' \emph{arXiv preprint arXiv:2004.14080}, 2020.

\bibitem{budzianowski2018large}
P.~Budzianowski, T.-H. Wen, B.-H. Tseng, I.~Casanueva, U.~Stefan, R.~Osman, and
  M.~Ga{\v{s}}i\'c, ``Multiwoz - a large-scale multi-domain wizard-of-oz
  dataset for task-oriented dialogue modelling,'' in \emph{Proceedings of the
  2018 Conference on Empirical Methods in Natural Language Processing (EMNLP)},
  2018.

\bibitem{eric2019multiwoz}
M.~Eric, R.~Goel, S.~Paul, A.~Sethi, S.~Agarwal, S.~Gao, and D.~Hakkani-Tur,
  ``Multiwoz 2.1: Multi-domain dialogue state corrections and state tracking
  baselines,'' \emph{arXiv preprint arXiv:1907.01669}, 2019.

\bibitem{WuTradeDST2019}
C.-S. Wu, A.~Madotto, E.~Hosseini-Asl, C.~Xiong, R.~Socher, and P.~Fung,
  ``Transferable multi-domain state generator for task-oriented dialogue
  systems,'' in \emph{Proceedings of the 57th Annual Meeting of the Association
  for Computational Linguistics (Volume 1: Long Papers)}.\hskip 1em plus 0.5em
  minus 0.4em\relax Association for Computational Linguistics, 2019.

\bibitem{chao2019bert}
G.-L. Chao and I.~Lane, ``Bert-dst: Scalable end-to-end dialogue state tracking
  with bidirectional encoder representations from transformer,'' \emph{arXiv
  preprint arXiv:1907.03040}, 2019.

\bibitem{heck2020trippy}
M.~Heck, C.~van Niekerk, N.~Lubis, C.~Geishauser, H.-C. Lin, M.~Moresi, and
  M.~Ga{\v{s}}i{\'c}, ``Trippy: A triple copy strategy for value independent
  neural dialog state tracking,'' \emph{arXiv preprint arXiv:2005.02877}, 2020.

\bibitem{kumar2020ma}
A.~Kumar, P.~Ku, A.~Goyal, A.~Metallinou, and D.~Hakkani-Tur, ``Ma-dst:
  Multi-attention-based scalable dialog state tracking,'' in \emph{Proceedings
  of the AAAI Conference on Artificial Intelligence}, vol.~34, no.~05, 2020,
  pp. 8107--8114.

\bibitem{trade-dst}
C.-S. Wu, A.~Madotto, E.~Hosseini-Asl, C.~Xiong, R.~Socher, and P.~Fung,
  ``Transferable multi-domain state generator for task-oriented dialogue
  systems,'' \emph{arXiv preprint arXiv:1905.08743}, 2019.

\bibitem{zhou2020multidomain}
L.~Zhou and K.~Small, ``Multi-domain dialogue state tracking as dynamic
  knowledge graph enhanced question answering,'' 2020.

\bibitem{zhang2019find}
J.-G. Zhang, K.~Hashimoto, C.-S. Wu, Y.~Wan, P.~S. Yu, R.~Socher, and C.~Xiong,
  ``Find or classify? dual strategy for slot-value predictions on multi-domain
  dialog state tracking,'' \emph{arXiv preprint arXiv:1910.03544}, 2019.

\bibitem{chen2020schema}
L.~Chen, B.~Lv, C.~Wang, S.~Zhu, B.~Tan, and K.~Yu, ``Schema-guided
  multi-domain dialogue state tracking with graph attention neural networks,''
  in \emph{Proceedings of the AAAI Conference on Artificial Intelligence},
  vol.~34, no.~05, 2020, pp. 7521--7528.

\bibitem{zhou2021dialogue}
J.~Zhou, H.~Wu, Z.~Lin, G.~Li, and Y.~Zhang, ``Dialogue state tracking with
  multi-level fusion of predicted dialogue states and conversations,'' 2021.

\bibitem{campagna2020zeroshot}
G.~Campagna, A.~Foryciarz, M.~Moradshahi, and M.~S. Lam, ``Zero-shot transfer
  learning with synthesized data for multi-domain dialogue state tracking,''
  2020.

\bibitem{su2021multitask}
Y.~Su, L.~Shu, E.~Mansimov, A.~Gupta, D.~Cai, Y.-A. Lai, and Y.~Zhang,
  ``Multi-task pre-training for plug-and-play task-oriented dialogue system,''
  2021.

\bibitem{hosseiniasl2020simple}
E.~Hosseini-Asl, B.~McCann, C.-S. Wu, S.~Yavuz, and R.~Socher, ``A simple
  language model for task-oriented dialogue,'' 2020.

\bibitem{lin2020mintl}
Z.~Lin, A.~Madotto, G.~I. Winata, and P.~Fung, ``Mintl: Minimalist transfer
  learning for task-oriented dialogue systems,'' \emph{arXiv preprint
  arXiv:2009.12005}, 2020.

\bibitem{peng2021soloist}
B.~Peng, C.~Li, J.~Li, S.~Shayandeh, L.~Liden, and J.~Gao, ``Soloist: Building
  task bots at scale with transfer learning and machine teaching,'' 2021.

\bibitem{raffel2019exploring}
C.~Raffel, N.~Shazeer, A.~Roberts, K.~Lee, S.~Narang, M.~Matena, Y.~Zhou,
  W.~Li, and P.~J. Liu, ``Exploring the limits of transfer learning with a
  unified text-to-text transformer,'' \emph{arXiv preprint arXiv:1910.10683},
  2019.

\bibitem{loshchilov2017decoupled}
I.~Loshchilov and F.~Hutter, ``Decoupled weight decay regularization,''
  \emph{arXiv preprint arXiv:1711.05101}, 2017.

\bibitem{wolf2020huggingfaces}
T.~Wolf, L.~Debut, V.~Sanh, J.~Chaumond, C.~Delangue, A.~Moi, P.~Cistac,
  T.~Rault, R.~Louf, M.~Funtowicz, J.~Davison, S.~Shleifer, P.~von Platen,
  C.~Ma, Y.~Jernite, J.~Plu, C.~Xu, T.~L. Scao, S.~Gugger, M.~Drame, Q.~Lhoest,
  and A.~M. Rush, ``Huggingface's transformers: State-of-the-art natural
  language processing,'' 2020.

\end{thebibliography}

\appendix

\section{Appendix}
\subsection{Parameter size}
\begin{table}[th]
\caption{Parameter size of baseline models and SF-DST.}
\label{tab:parameters}
\centering
\begin{tabular}{lc}
\toprule
Models       & \#Parameter \\ \midrule
MinTL \cite{lin2020mintl}  & 400M        \\
GPT2QA \cite{li2021zeroshot}       & 355M       \\
SST-2 \cite{chen2020schema}        & 110M        \\
TripPy \cite{heck2020trippy}        & 110M        \\
SimpleTOD \cite{hosseiniasl2020simple}    & 110M        \\
STARC \cite{gao2020machine}        & 110M        \\
SOLOIST\cite{peng2021soloist}      & 110M        \\
PPTOD$_{small}$ \cite{su2021multitask} & 60M         \\ \midrule
SF-DST(ours)         & 60M         \\ \bottomrule
\end{tabular}
\end{table}

\subsection{Detailed implementation}
We implement SF-DST using T5-small \cite{raffel2019exploring}, which has six encoder/decoder layers, and the hidden model has size 512. All models are trained using an NVIDIA A5000 GPU for five epochs with early stopping. For optimization, we use AdaFactor \cite{loshchilov2017decoupled}. The batch size is 16 in the few-shot setting and 32 in the supervised setting. We implement T5-small based on Huggingface Library \cite{wolf2020huggingfaces}.

\end{document}